\documentclass[pmlr]{jmlr}

\usepackage{longtable}
\usepackage{amsmath} 

\usepackage{booktabs}
\usepackage[load-configurations=version-1]{siunitx} 

\theorembodyfont{\upshape}
\theoremheaderfont{\scshape}
\theorempostheader{:}
\theoremsep{\newline}

\jmlrvolume{1}
\jmlryear{2026}
\jmlrworkshop{Impactful and Responsible AI Systems for Education (IRAISE)}
\usepackage{amsmath}
\usepackage{hyperref}
\usepackage{cleveref}   

\title[Uncertainty for KT Models]{Knowing When to Defer: Selective Prediction for Responsible Knowledge Tracing}

\newcommand{\eqcontrib}{\textsuperscript{\dag}}
\author{\Name{Joshua Mitton\eqcontrib} \Email{Eedi}\\
 \Name{Prarthana Bhattacharyya\eqcontrib} \Email{Eedi}\\
 \Name{Ralph Abboud} \Email{Renaissance Philanthropy}\\
 \Name{Simon Woodhead} \Email{Eedi}}

\begin{document}

\maketitle
\makeatletter
\long\def\@makefntext#1{\noindent #1}
\makeatother
{\let\thefootnote\relax\footnotetext{\textsuperscript{\dag} Equal contribution.}}
\makeatletter
\long\def\@makefntext#1{\@setpar{\@@par\@tempdima \hsize
  \advance\@tempdima-15pt\parshape \@ne 15pt \@tempdima}\par
  \parindent 2em\noindent \hbox to \z@{\hss{\@thefnmark}. \hfil}#1}
\makeatother
\setcounter{footnote}{0}
\thispagestyle{empty} 
\begin{abstract}
Research on Knowledge Tracing (KT) models traditionally focuses on improving predictive accuracy. However, responsible real-world deployment requires models to know when to defer uncertain predictions to a human teacher. We introduce an intrinsic selective prediction layer for existing KT models using Monte Carlo Dropout (MC-Dropout) to quantify uncertainty. We evaluate this approach across three architectures (DKT, SAKT, and AKT) using the Eedi mathematics dataset. Abstaining on the 20\% most uncertain predictions lifts accuracy by 2.3 to 3.0 percentage points, AUC by 1.9 to 2.4 percentage points and F1 by 1.4 to 4.3 percentage points without any retraining. This abstention strategy is highly targeted: the deferred set exhibits 1.45 to 1.60 times the error rate of the kept set. Furthermore, this targeting holds within every question-difficulty quartile and remains fair across student-ability levels. Importantly, MC-Dropout variance gives roughly five times the AUC lift of a calibrated two-parameter logistic (2PL) Item Response Theory (IRT) baseline as a selective-prediction signal. A variance decomposition of the model's epistemic uncertainty (BALD) reveals that the entire classical psychometric stack, comprising question difficulty, student ability, IRT-style outcome ambiguity, and historical curriculum coverage, explains less than 4\% of the signal under linear modeling and at most 23\% even with a non-linear regressor. This leaves 77\% to 90\% as architecture-specific epistemic content that MC-Dropout surfaces and simpler proxies cannot recover. Selective prediction with model-native epistemic uncertainty is therefore a necessary component of responsible KT deployment, complementary to subgroup-fairness audits and downstream classroom evaluation rather than a substitute for them.

\end{abstract}
\begin{keywords}
Knowledge tracing, Uncertainty, MC dropout.
\end{keywords}

\section{Introduction}
\label{sec:intro}
Knowledge Tracing (KT) models are widely used to predict how a student will respond to future questions based on their past interactions. These models have evolved significantly over time, transitioning from early Bayesian approaches \cite{pardos2010modeling,yudelson2013individualized} to deep learning architectures built on LSTMs \cite{DKT}, self-attention \cite{SAKT,AKT}, and graphs \cite{yang2020gikt}, and more recently to frameworks leveraging Large Language Models \cite{wang2025llm}. Despite these architectural leaps, actual performance gains in accuracy are often marginal \cite{ozyurt2024automated}. More importantly, existing KT models typically struggle with low specificity, which allows incorrect student answers to slip by undetected during real-world deployment \cite{yamkovenko2025practical}.

We address this critical gap by treating responsible KT deployment as a selective prediction problem. To achieve this, we introduce an intrinsic selective prediction layer designed to integrate directly with existing KT models. The proposed system calculates its own uncertainty, abstains from making the most unreliable predictions, and gracefully defers those specific cases to a human teacher. We base our analysis on data collected from a live educational setting through the Eedi mathematics dataset. By applying Monte Carlo Dropout~\cite{mcdropout} to quantify uncertainty across three distinct KT architectures (DKT, SAKT, and AKT), we offer four primary contributions:

\begin{itemize}
    \item We demonstrate that selective prediction improves accuracy, AUC, and F1 scores across all three architectures without requiring any retraining. This abstention is highly targeted, with deferred predictions showing an error rate 1.45 to 1.60 times higher than the kept predictions. The targeting holds true within every quartile of question difficulty and remains equitable across different student ability levels.
    
    \item We show that MC-Dropout variance outperforms a calibrated two-parameter logistic (2PL) Item Response Theory (IRT) baseline on the same selective prediction task, giving roughly five times the AUC lift at 80 percent coverage.
    
    \item We use variance decomposition of the model's epistemic uncertainty (BALD) to show that the entire classical psychometric stack (question difficulty, student ability, IRT-style outcome ambiguity, and historical curriculum coverage at four granularities) accounts for less than 4 percent of the signal under linear modeling and at most 23 percent even with a non-linear regressor (5-fold cross-validation, random forest). The remaining 77 to 90 percent is architecture-specific epistemic content: the model's uncertainty over its own learned representations of a student's unique sequential trajectory, which simpler models cannot capture even with the freedom of non-linear modeling.
    
    \item Finally, we establish that responsible KT deployment must rely on the model's intrinsic uncertainty signal, rather than falling back on simple psychometric proxies or heuristic coverage rules.
\end{itemize}
\section{Related Works}

\paragraph{Deep Knowledge Tracing Models}
Knowledge Tracing (KT) models the temporal dynamics of student learning by predicting knowledge states based on sequences of interactions with educational content \cite{Corbett2005KnowledgeTM}.
Various machine learning methods, including deep neural networks, have been proposed to address this sequential prediction task.
Among these, deep sequential models adopt auto-regressive architectures to capture the progression of student interactions.
Deep Knowledge Tracing (DKT) \cite{DKT} was one of the first to use an LSTM to estimate students' knowledge states over time.
Subsequent work, such as KQN \cite{kqn}, enhanced KT by encoding student learning activities into knowledge state and skill vectors and modeling their interactions with the dot product.
Other lines of work introduce memory-augmented (DKVMN~\cite{DKVMN}) and graph-based representations of knowledge components (e.g., GIKT \cite{gikt,liugraph2019}).

\paragraph{Attention-Based Architectures}
However, recurrent and attention-based models remain the dominant deployed families and are the focus of this study.
Attention-based models leverage attention mechanisms to dynamically weigh past interactions.
SAKT \cite{SAKT} introduced a self-attention architecture to model the relevance of historical responses.
AKT \cite{AKT} introduced monotonic attention modules to model forgetting behavior.

\paragraph{Uncertainty Modeling in KT}
Despite these advances, uncertainty modeling in KT remains underexplored.
Most existing models produce point predictions without quantifying the confidence in knowledge estimates.
This is an important limitation for real-world educational applications.
UKT \cite{UKT} is the only other work to the best of our knowledge that introduces a KT model to capture uncertainty in student learning by representing knowledge states as stochastic distributions and modeling their transitions with a Wasserstein self-attention mechanism.
While UKT provides an approach to modeling uncertainty, it introduces an entirely new architecture, which limits its applicability in real-world settings where existing KT models are already deployed.
Adding uncertainty quantification on top of such models remains non-trivial with this method.

\paragraph{Learning to Defer}
Outside the KT literature, the broader machine-learning community has formalized the abstention-and-defer paradigm we operationalize here under the umbrella of learning to defer ~\citep{madras2018predict,mozannar2020consistent}.
In this paradigm, a classifier explicitly chooses between predicting and routing to a human expert.
Our contribution adapts this approach to a deployed KT setting using model-native epistemic uncertainty as the routing signal rather than a learned deferral head.
\section{Predictive Uncertainty for Knowledge Tracing Models}
\subsection{Knowledge Tracing Models}
We adopt the standard formulation of knowledge tracing \cite{DKT}, which predicts a student’s performance on a future exercise based on their sequence of past interactions. Formally, we model a student's interaction history up to time $t$ as a sequence of exercises, $\{e_i\}_{i=1}^{t}$. Each exercise is represented as a three-tuple, $e_i = (q_i, c_i, r_i)$. Here, $q_i$ denotes the unique question identifier, $c_i$ encapsulates the pedagogical features associated with $q_i$ (such as knowledge component identifiers), and $r_i \in \{\text{Correct}, \text{Incorrect}\}$ indicates the binary correctness of the student's response.

Given this historical sequence and the context of an upcoming question, the KT model $F_\theta$ predicts the probability of a correct outcome $\hat{r}_{t+1}$ for the next exercise. Formally, this prediction is defined as:
$$
\hat{r}_{t+1} = F_\theta\left(q_{t+1}, c_{t+1}, \{e_i\}_{i=1}^{t}\right)
$$

\subsubsection{Baseline KT Models}
To benchmark KT approaches for mathematics education, we consider three models as instantiations of \( F_\theta \): Deep Knowledge Tracing (DKT)~\cite{DKT}, Self-Attentive Knowledge Tracing (SAKT)~\cite{SAKT}, and Attentive Knowledge Tracing (AKT)~\cite{AKT}. DKT models student learning with recurrent neural networks (RNNs). It takes as input the sequence of question identifiers and responses $(q_i, r_i)$ and, conditioned on the next question $q_{t+1}$, predicts the probability of a correct response $\hat{r}_{t+1}$ by capturing temporal dependencies in the interaction history. SAKT replaces recurrence with a self-attention mechanism, which allows the model to selectively attend to relevant past interactions when predicting performance on the next question. AKT extends this approach by introducing contextual attention that considers both the similarity between questions and an exponential decay to model student forgetting. Together these three architectures span the dominant families of modern KT models (recurrent and attention-based), letting us test whether our claims about uncertainty are architecture-invariant. These models are typically evaluated using point-estimate metrics such as accuracy, F1 or AUC, which quantify predictive performance but provide no measure of model uncertainty. This gap motivates the Bayesian formulation introduced next.
\subsection{MCDropout for KT Uncertainty}
We estimate predictive uncertainty using Monte Carlo Dropout (MC Dropout) \cite{mcdropout}.
At inference, dropout remains active and predictions are averaged over $M$ stochastic forward passes.
Following the information-theoretic approach based on Shannon entropy~\cite{CoverandThomas2006},
we quantify total uncertainty by the entropy of the aggregated predictive distribution:
\[
H\big(p(r_{t+1}\mid q_{t+1}, \mathcal{H}_t)\big)
= -\sum_{k} \bar{p}_k \log \bar{p}_k,
\quad
\bar{p}_k \;=\; \frac{1}{M}\sum_{m=1}^{M} p_k^{(m)}.
\]
where $\bar{p}_k$ denotes the mean predictive probability of class $k$ estimated via MCDropout. Here, $\mathcal{H}_t = \{ e_i \}_{i=1}^t$ represents the interaction history of a student up to time $t$,
with each exercise $e_i = (q_i, c_i, r_i)$ defined by the question identifier $q_i$, its associated features or knowledge components $c_i$, and the observed response $r_i$. Higher entropy corresponds to greater uncertainty in the model's prediction, whereas lower entropy reflects more confident predictions. We additionally use the variance and standard deviation of the predictions across MC samples as complementary uncertainty measures to operationalize the selective prediction layer (\Cref{app:model_eval}).

\section{Experiments and Results}

\subsection{Architectures and dataset}
We implemented our selective prediction approach across three KT architectures: DKT~\cite{DKT}, SAKT~\cite{SAKT}, and AKT~\cite{AKT}. These models represent the dominant recurrent and attention-based families in current KT literature. 

The training dataset consists of student responses to mathematics questions. To standardize the context, we limited the interaction history to 100 question-answer pairs per student. Within this platform, students complete questions in five-question quiz sessions that progressively increase in difficulty. Consequently, the 100-interaction history typically reflects a student's performance across 20 distinct quizzes. Further dataset details are provided in Section~\ref{app:dataset_details}.

\subsection{Qualitative uncertainty patterns}
Across all three architectures, a clear qualitative pattern emerges. Misclassified predictions concentrate at higher MC-Dropout entropy and standard deviations than correctly classified ones. The uncertainty signal varies systematically with the within-quiz difficulty progression. \Cref{sec:uncertainty_app} provides the distribution and trajectory plots that establish this behavior. While a monotonic correlation between uncertainty and model error is encouraging, the critical challenge is translating this signal into a practical deployment layer.

\subsection{Selective Prediction: Translating Uncertainty into a Deployment Layer}
\label{sec:selpred}

\begin{figure}[t]
  \centering
  \includegraphics[width=\linewidth]{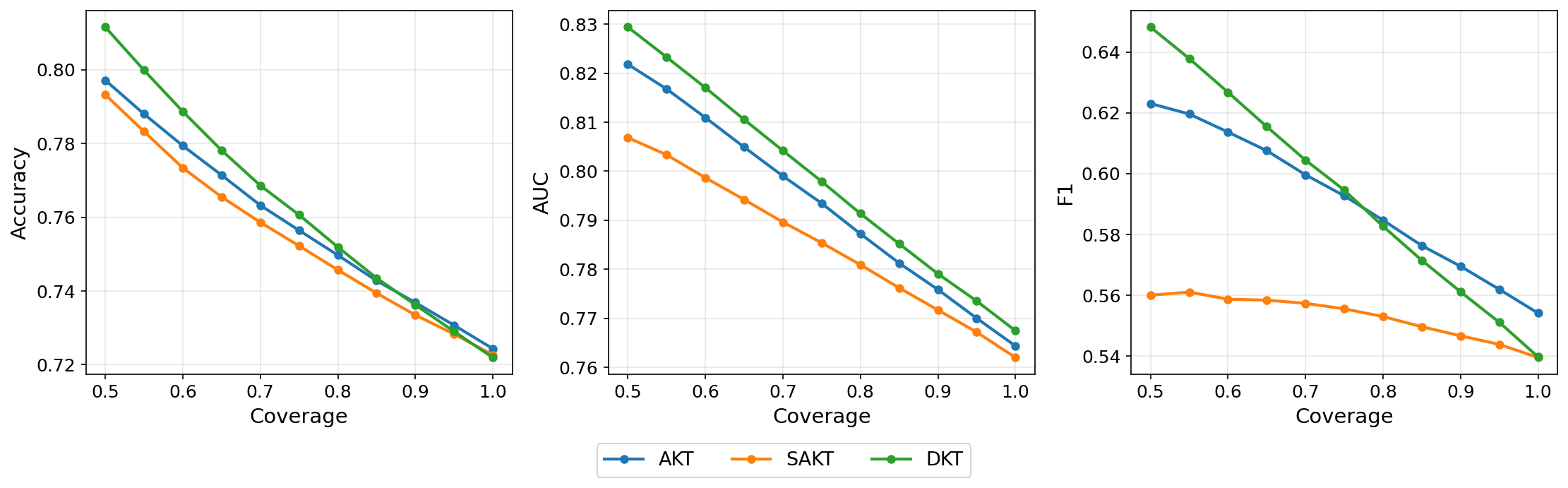}
  \caption{Selective prediction with MC-Dropout variance: every metric improves monotonically as we defer the most-uncertain cases, on all three architectures.}
  \label{fig:selective_curves}
\end{figure}

To operationalize the uncertainty signal, we sort predictions by their MC-Dropout variance in ascending order and abstain from the most uncertain fraction to achieve a target coverage rate $c$. \Cref{fig:selective_curves} plots accuracy, AUC, and F1 scores against coverage for the binary correctness target. Across all three architectures, every metric improves monotonically as coverage decreases, achieving these gains entirely without retraining. At an operating point of $c=0.80$ (where 20\% of cases are deferred to a teacher), MC-Dropout variance increases accuracy by 2.3 to 3.0 percentage points (pp), AUC by 1.9 to 2.4 pp, and F1 by 1.4 to 4.3 pp. All lifts are statistically significant, as the 95\% bootstrap confidence intervals (200 resamples) never cross zero. These intervals are exceptionally tight across all architectures, with margins of error of just $\pm 0.13$\,pp for the change in accuracy, $\pm 0.10$\,pp for AUC, and $\pm 0.20$\,pp for F1.

\begin{figure}[t]
  \centering
  \includegraphics[width=\linewidth]{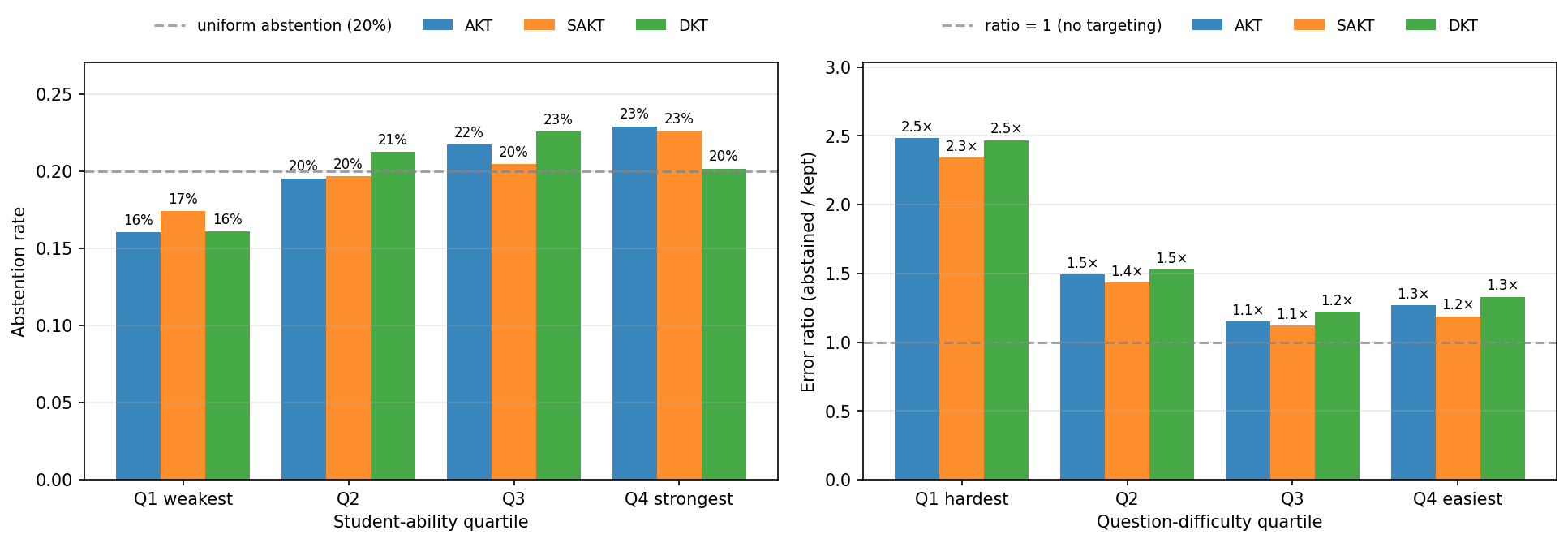}
  \caption{Selective prediction at $c=0.80$ using the MC-Dropout variance signal is both fair and targeted. \textbf{(a)} Abstention rate by student-ability quartile: all four quartiles fall close to the 20\% uniform baseline, and the weakest students are abstained on \emph{less}, not more, than the rest. \textbf{(b)} Error ratio (abstained / kept) within each question-difficulty quartile: every (model $\times$ quartile) cell is above 1.0, with the strongest targeting on the hardest questions (2.3--2.5$\times$).}
  \label{fig:targeting_fairness}
\end{figure}

\paragraph{Targeting and fairness.}
This abstention mechanism is highly targeted. At 80\% coverage, the error rate of the deferred set is 1.45 to 1.60 times higher than that of the kept set across the three models. To ensure this effect is not simply an artifact of question difficulty, we recalculated the error ratio within each question-difficulty quartile (\Cref{fig:targeting_fairness}b). The targeting holds true within every stratum: every combination of model and question-difficulty quartile maintains an error ratio greater than 1.0, with the strongest targeting occurring on the hardest questions (2.3 to 2.5 times higher error rates across all three architectures). Equally important, the abstention is fair across student ability. \Cref{fig:targeting_fairness}a shows that the abstention rate ranges tightly between 16\% and 23\% across all four student-ability quartiles, close to the 20\% uniform baseline. The weakest students (Q1) actually receive the \emph{lowest} abstention rate (16--17\%), ruling out the failure mode in which a model defers low-ability students to teachers by default.

\subsection{Comparison to a calibrated IRT baseline}
\begin{figure}[t]
  \centering
  \includegraphics[width=\linewidth]{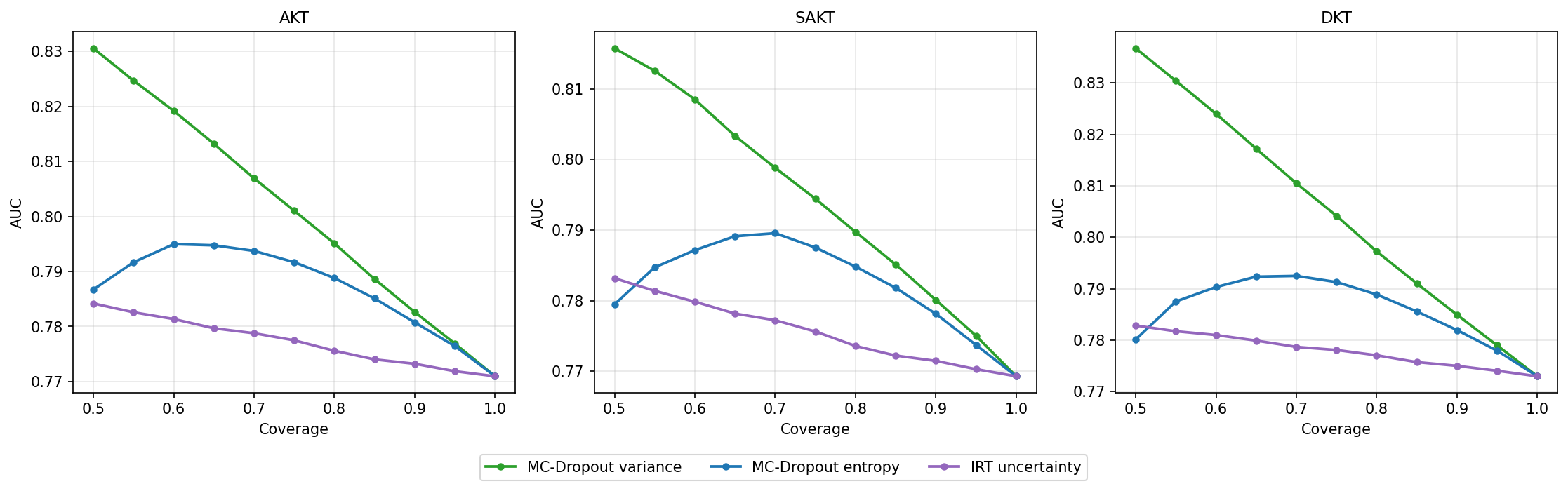}
  \caption{MC-Dropout variance dominates a calibrated 2PL IRT baseline as a selective-prediction signal across all three architectures (last-70 target subset).}
  \label{fig:irt_baseline}
\end{figure}

A simpler alternative to our approach would be to abstain on predictions that a calibrated two-parameter logistic (2PL) Item Response Theory (IRT) \citep{Lord1966SomeLT} model identifies as ambiguous. The validation cohort is held out by student, so a pre-fit IRT model trained on the training split has no ability parameter for these previously-unseen learners. We therefore fit each validation student's ability $\theta_u$ from a held-out portion of their own response history. We use each user's first 30 questions as the fitting window, estimating $\theta_u$ via maximum likelihood with item parameters $a_q, b_q$ taken from an IRT table pre-fitted on the training cohort over the question pool. The IRT abstention signal $1-2|p_{\mathrm{IRT}}-0.5|$ is then evaluated on their remaining 70 targets, and the MC-Dropout comparison is restricted to the same matched subset ($\sim$110{,}000 targets per model).\footnote{The headline lifts hold on this matched subset: MC-Dropout variance gives $\Delta\text{acc}@80=+2.4$--$3.0$\,pp, $\Delta\text{AUC}@80=+2.0$--$2.4$\,pp, $\Delta\text{F1}@80=+1.6$--$4.6$\,pp, within $0.3$\,pp of the full-sequence numbers.} \Cref{fig:irt_baseline} compares MC-Dropout variance, MC-Dropout entropy, and the IRT uncertainty baseline on this matched subset. MC-Dropout variance yields an AUC lift of 2.0 to 2.4 percentage points at 80\% coverage. In contrast, the IRT baseline provides a lift of only 0.41 to 0.46 percentage points, roughly five times smaller. This demonstrates that MC-Dropout is not merely a complex re-implementation of IRT-style confidence. Instead, its stochastic forward passes uncover genuine epistemic information that a calibrated psychometric model fundamentally misses. We note that 2PL IRT is one specific baseline, bounded by its two-parameter representational capacity. Alternative selective-prediction baselines such as a temperature-scaled softmax of the deployed model itself~\citep{guo2017calibration} or a deep ensemble of independently trained checkpoints~\citep{lakshminarayanan2017simple} would test the irreducibility claim more strongly, and remain a natural follow-up.

\subsection{Drivers of predictive uncertainty}
To determine what exactly this uncertainty tracks, we ask whether MC-Dropout's signal is recoverable from cheaper, model-free features of (student, question) pairs. We regress the model's \emph{epistemic uncertainty} (BALD: total entropy minus expected aleatoric entropy across MC samples) on a sequence of nested predictor sets and report cumulative $R^2$ at each step. Predictors are added in order: question difficulty (empirical correctness rate), student ability (per-user accuracy), IRT-style outcome ambiguity ($H(p_{\mathrm{IRT}})$), and curriculum coverage at four granularities (binary indicators for whether the target's subject, topic, subtopic, or construct appears in the student's first-30 context). The full procedure is detailed in Appendix~\ref{sec:variance_decomp_app}.

Across all three architectures, every classical psychometric factor yields negligible explanatory power. Question difficulty alone explains 0.4\% to 1.5\% of the variance. Adding student ability brings the cumulative total to 0.8\% to 1.8\%. IRT outcome ambiguity raises it to between 1.1\% and 3.7\%. Adding all four levels of curriculum coverage contributes a further 0.0\% to 0.2\%, leaving the cumulative linear-explained variance at most 3.8\% across architectures (\Cref{fig:what_drives}, Appendix; see Appendix~\ref{app:curriculum} for overlap rates). To bound how much of this gap can be closed by a stronger model class, we re-fit the same predictor stack with a non-linear regressor (5-fold cross-validated random forest). Explained variance rises to 9.8\% (SAKT), 10.5\% (AKT), and 23.2\% (DKT), still leaving \textbf{76.8\% to 90.2\%} unexplained.

This residual, dominant even under non-linear modeling, is exactly what MC-Dropout captures: the architecture's native uncertainty over its own learned, non-linear representations of a student's unique sequential trajectory. Heuristic coverage-tracking rules and 2PL-confidence calibrators are structurally incapable of recovering it. By construction they only have access to a small set of psychometric features (ability, difficulty, outcome ambiguity). Even given the freedom of a non-linear classifier, those features leave the majority of the signal architecture-specific. A model-native probe such as MC-Dropout is therefore necessary to surface this hidden uncertainty and enable responsible deployment.
\section{Conclusion}

Selective prediction with model-native epistemic uncertainty is a necessary component of responsible KT deployment. It is complementary to subgroup fairness audits, calibration analysis, and downstream classroom evaluation rather than a replacement for them. For confidence-aware systems to operate safely in real classrooms, they must reliably distinguish between safe and risky predictions using signals derived directly from their own internal representations. We have demonstrated that cheaper heuristic alternatives such as population statistics, IRT baselines, or curriculum coverage miss the vast majority of this per-prediction reliability signal by construction.

The model's native MC-Dropout epistemic uncertainty successfully targets and defers high-risk predictions across DKT, SAKT, and AKT architectures. The deferred set is also fair across student-ability and question difficulty quartiles. As a concrete deployment example, an Eedi-style next-question-recommendation pipeline could use the variance signal as a gate. Low-variance predictions would feed directly into the recommender's ranking. High-variance predictions would trigger a fallback (such as a teacher review queue, an easier diagnostic question, or a mastery quiz on the relevant construct), rather than committing to a low-confidence next-question pick. The architectural and analytical work in this paper grounds that integration. The operational work of validating it against student-outcome metrics in a real classroom remains future work.

\paragraph{Limitations.}
Our claims are bounded by four constraints. First, the headline lifts (2.3 to 3.0\,pp accuracy at $c=0.80$) are improvements on validation-set predictive metrics, not on student learning outcomes or teacher cognitive load. We have not run a classroom A/B test or simulated deployment, so the responsibility argument is conditional on those follow-ups. Second, our analysis uses a single dataset (Eedi mathematics responses), and the architecture-specific epistemic dominance has not been validated on other KT datasets. Third, the IRT comparison uses a 2-parameter logistic model fit on 30 questions per validation student. Alternative selective-prediction baselines such as a temperature-scaled softmax of the deployed model or a deep ensemble are not evaluated and remain promising future comparisons. Fourth, for the IRT baseline, we do not address cold-start: students with fewer than 30 prior responses fall outside our $\theta$-fitting window.

\paragraph{AI-tool disclosure.}
AI assistants were used during the preparation of this manuscript for code review, copy-editing, and figure-style suggestions. All experimental design, model training, analysis, methodological decisions, and conclusions are the authors' own.
\bibliography{references}

\newpage
\appendix
\section{Experimental Setup}
\subsection{Dataset Details}
\label{app:dataset_details}
We use a filtered subset of question-response logs from Eedi, an online mathematics learning platform. An interaction is a single question–response pair. \\
The corpus contains \(4{,}257\) unique questions. Each question is multi-choice, and has question-ID and question-text associated with it. For each question, there are four options to choose from:  \{A, B, C, D\}. We have the labels for which of those options is correct, and the other three options are carefully curated distractors tied to a mathematics misconception. The additional features a question contains includes construct-text, construct-ID, explanation-text and misconception-text. A construct-text is the most granular level of knowledge related to question, for example, `Construct 1: Find 10 less than a given number'. A misconception-text describes a cognitive error per distractor option which leads to mistakes. It is generic, unlike an explanation-text per option, which tends to be specific to the question. \\
We first filter students to those with at least 100 recorded responses and then retain the final 100 responses per student to standardize sequence length. The resulting training split contains \(\,11{,}994\,\) students (average \(100\) responses; \(1{,}199{,}266\) total interactions), and the validation split contains \(\,1{,}493\,\) students (average \(100\) responses; \(149{,}278\) total interactions) as shown in Table~\ref{tab:data_stats}. 
All splits are student-disjoint: validation students do not appear in training. 
During evaluation, models predict each of the \(100\) student responses in sequence with access only to prior interactions (causal/left-to-right), never the current or future response.

\begin{table}[h]
\centering
\caption{Dataset statistics after filtering. Interactions are question–response pairs. Splits are disjoint by student.}
\label{tab:data_stats}
\begin{tabular}{lrrr}
\toprule
Split & \# Students & Avg. responses & Total interactions \\
\midrule
Train & 11{,}994 & 100 & 1{,}199{,}266 \\
Val   & 1{,}493  & 100 &   149{,}278 \\
\bottomrule
\end{tabular}
\end{table}

\subsection{Curriculum Hierarchy and Context-Target Overlap}
\label{app:curriculum}

Each Eedi question is tagged with four nested levels of curriculum metadata. Ranging from coarsest to finest, these levels are \textit{subject}, \textit{topic}, \textit{subtopic}, and \textit{construct}. The curriculum forms a strict hierarchy, meaning every question belongs to exactly one category at each of these four levels. For example, question 31769 in our dataset falls under the subject `Number', the topic `Proportion', and the subtopic `Direct Proportion'. At the most granular level, its construct is defined as `Use direct proportion to solve non-unitary missing amounts in problems (e.g. recipes)'. 

\Cref{tab:curriculum_levels} details the total cardinality of each curriculum level. It also reports the average context-target overlap rate. We define this overlap as the fraction of a student's final 70 prediction targets that share a specific curriculum tag with at least one question from their initial 30-question context.

As expected, this coverage drops sharply as granularity increases. Because the dataset contains only six broad subjects, the majority of target questions share a subject with the historical context, while only about 6\% of targets share a specific construct. The variance decomposition presented in Section~\ref{sec:selpred} explicitly controls for coverage across \emph{all four} of these levels simultaneously. Consequently, our conclusion that curriculum coverage fails to drive uncertainty remains robust, regardless of the specific granularity at which a deployment system might track student progress.

\begin{table}[h]
\centering
\caption{Curriculum hierarchy: cardinality and average context-target overlap rate (across the three architectures, on the last-70 prediction subset given a first-30 context).}
\label{tab:curriculum_levels}
\begin{tabular}{lrr}
\toprule
Level & \# unique values & \% targets where level appears in context \\
\midrule
Subject   & 6      & 81.4\% \\
Topic     & 50     & 29.1\% \\
Subtopic  & 212    & 14.1\% \\
Construct & 1{,}463 & 6.2\%  \\
\bottomrule
\end{tabular}
\end{table}

\subsection{Model Training}
We train three knowledge tracing models: DKT \cite{DKT}, SAKT \cite{SAKT}, and AKT \cite{AKT}, under a unified setup. Model sizes are summarized in \Cref{tab:model_params}. AKT is the largest (3.3~M parameters), DKT is the most compact (1.2~M). Unless otherwise noted, all models share the global training configuration in \Cref{tab:global_hparams}: Adam optimizer, learning rate \(3\times10^{-4}\), cross-entropy loss, batch size 64, and 100 epochs, with a linear warmup followed by a cosine scheduler. Model-specific hyperparameters are listed in \Cref{tab:model_hparams}. We fix embedding and hidden dimensions to 128 and use one layer across models, with dropout tuned per architecture (0.5 for SAKT, 0.2 for AKT and DKT). AKT uses the \texttt{kq\_same} setting. All other details are held constant to enable a fair comparison across architectures.

\paragraph{Reproducibility.}
MC-Dropout inference uses $M = 100$ stochastic forward passes per prediction with dropout active at the rates listed above; aleatoric and epistemic components are computed via the BALD decomposition (Section~\ref{sec:variance_decomp_app}). All training and inference uses a single fixed random seed; multi-seed robustness is not reported in this version. 
\begin{table}[htb]
\label{model params and memory}
\centering
\caption{Parameter counts and estimated memory footprint for each model. We report total trainable parameters (in millions) and the estimated model size (in MB).}
\label{tab:model_params}
\begin{tabular}{lrr}
\toprule
Model & Total params (M) & Est. size (MB) \\
\midrule
DKT \cite{DKT}             & 1.2 & 4.890 \\
SAKT \cite{SAKT}           & 1.7 & 6.990 \\
AKT  \cite{AKT}           & 3.3 & 13.020 \\
\bottomrule
\end{tabular}
\end{table}

\begin{table}[htb]
\centering
\caption{Global training hyper-parameters used across all models.}
\label{tab:global_hparams}
\begin{tabular}{ll}
\toprule
Setting     & Value \\
\midrule
Learning rate (LR) & 0.0003 \\
Optimizer & Adam \\
Loss & Cross-Entropy \\
Batch size & 64 \\
Epochs & 100 \\
Scheduler & Linear (warmup), cosine (training) \\
\bottomrule
\end{tabular}
\end{table}

\begin{table}[htb]
\centering
\caption{Model-specific hyper-parameters.}
\label{tab:model_hparams}
\small
\begin{tabular}{lccc}
\toprule
 & DKT & SAKT & AKT \\
\midrule
Dropout & 0.2 & 0.5 & 0.2 \\
Embedding dim & 128 & 128 & 128 \\
Hidden dim & 128 & 128 & 128 \\
{kq\_same} & - & - & True \\
Num layers & 1 & 1 & 1 \\
\bottomrule
\end{tabular}
\end{table}


\subsection{Model Evaluation}
\label{app:model_eval}
We use standard deviation of the model's predictions over Monte-Carlo samples as a metric for predictive uncertainty. The standard deviation per question per student is calculated as:

\[
\mathrm{\sigma}\big(p\big) = \mathbb{E}_{k} \big[ \sigma(p_k) \big], \quad \mathrm{\sigma}\big(p_k\big) = \sqrt{\tfrac{1}{M}\sum_{m=1}^{M} \left[ \big(p_{k}^{(m)} - \mu\big)^{2} \right]}
\]
where $k$ denotes the class, $p_{k}^{(m)}$ denotes the predicted probability of sample $m$, $\mu$ is the mean predicted probability, and $\mathrm{\sigma}\big(p_k\big)$ is the standard deviation of the model's predictions over $M$ Monte-Carlo samples.

\section{Model Quantitative Predictive Accuracy}
On the validation split, all three architectures attain similar baseline accuracy on the binary correctness prediction task (see \Cref{tab:kt_results}): AKT (\(72.44\%\)), SAKT (\(72.27\%\)), and DKT (\(72.20\%\)). F1 also clusters closely: AKT yields the strongest F1 (\(55.42\%\)), followed by DKT (\(53.98\%\)) and SAKT (\(53.96\%\)). AUC is similarly tight, ranging from 76.20\% (SAKT) to 76.75\% (DKT). These results suggest that while overall predictive performance is comparable across architectures, the more informative deployment-relevant metric is the quality of each model's per-prediction uncertainty signal, which we evaluate via selective prediction in Section~\ref{sec:selpred}.
\begin{table}[htb]
\centering
\caption{Baseline binary correctness prediction on the validation split (no abstention).}
\label{tab:kt_results}
\begin{tabular}{lcccc}
\toprule
Model  & Accuracy & F1 & AUC \\
\midrule
DKT \cite{DKT}   & 72.20          & 53.98          & \textbf{76.75} \\
SAKT \cite{SAKT} & 72.27          & 53.96          & 76.20          \\
AKT \cite{AKT}   & \textbf{72.44} & \textbf{55.42} & 76.43          \\
\bottomrule
\end{tabular}
\end{table}


\section{Uncertainty Analysis}
\label{sec:uncertainty_app}

\subsection{Uncertainty by Prediction Correctness}
\label{app:uncertainty_correctness}

\begin{figure}[ht]
  \centering
  \includegraphics[width=\linewidth]{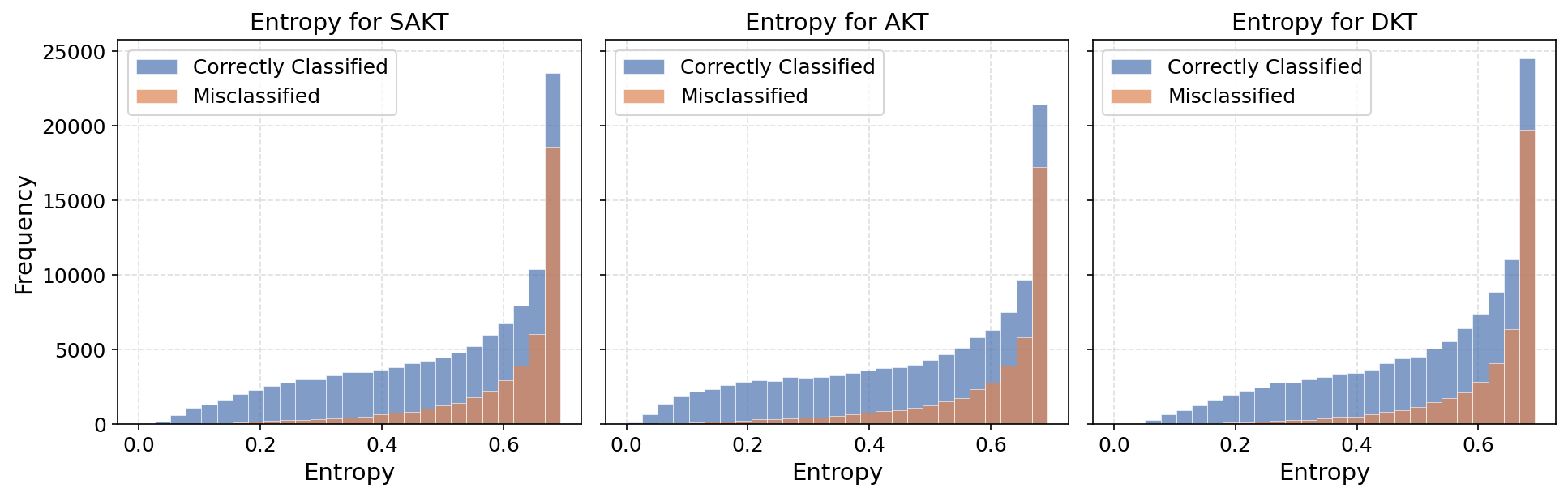}
  \caption{Total entropy distribution of each model's predictions as a function of their correctness. Misclassified data points clearly demonstrate a higher level of uncertainty.}
  \label{fig:total_entropy}
\end{figure}

The mean total entropy is consistently larger when the model's prediction is incorrect (\Cref{fig:total_entropy}). This confirms that our uncertainty measure successfully flags specific model errors: which is precisely the operational relationship that selective prediction (Section~\ref{sec:selpred}) exploits. 

\begin{figure}[ht]
    \begin{minipage}[t]{0.49\linewidth}
        \centering
        \includegraphics[width=\linewidth]{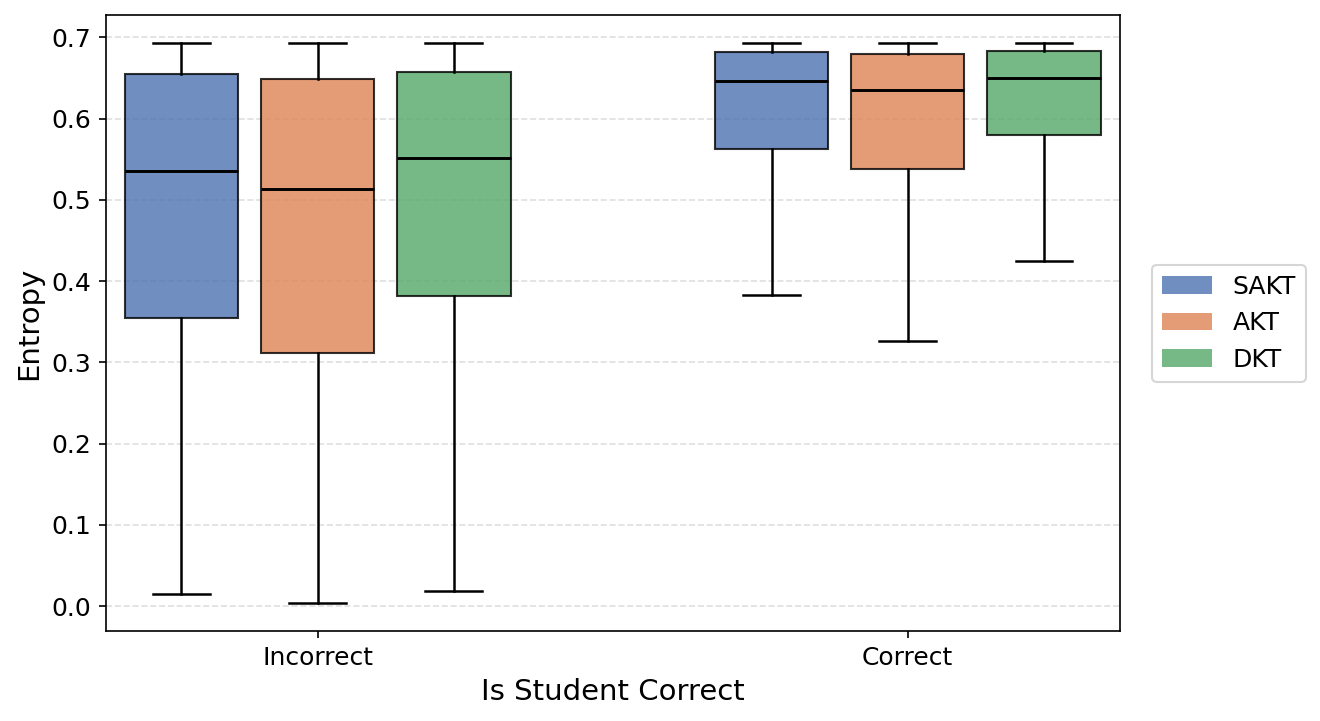}
        \caption{Box plot of total entropy of each model's predictions split into whether the student chose a correct response or an incorrect response.}
        \label{fig:entropy_box_plot}
    \end{minipage}
    \hfill
    \begin{minipage}[t]{0.49\linewidth}
        \centering
        \includegraphics[width=\linewidth]{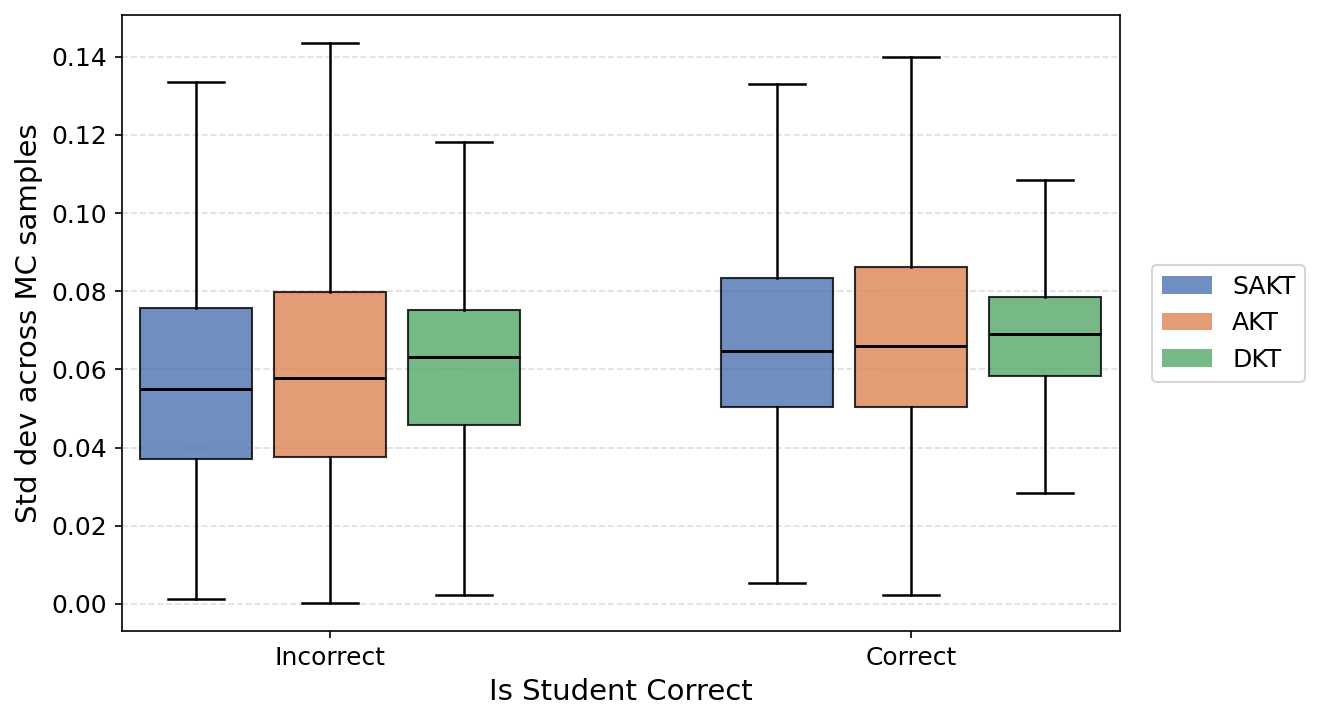}
        \caption{Box plot of each model's prediction standard deviation split into whether the student chose a correct response or an incorrect response.}
        \label{fig:std_box_plot}
    \end{minipage}
\end{figure}

However, when predictions are split by the \emph{student's} outcome rather than the \emph{model's} correctness, this relationship inverts. As shown in \Cref{fig:entropy_box_plot} and \Cref{fig:std_box_plot}, both median entropy and prediction standard deviation are slightly higher on student-correct cases than on student-incorrect cases. This is an artifact of binary correctness prediction under class imbalance. On hard questions where a student is likely to fail, the model can confidently predict failure, clustering low-entropy mass on the student-incorrect side. On easier questions where the student succeeds, the model commits to a moderately confident correct prediction, pushing the entropy closer to the binary maximum. 

Therefore, splitting by student outcome is not the appropriate diagnostic for model reliability; the model-correctness split (\Cref{fig:total_entropy}) is the true indicator. Finally, we note that a perfectly calibrated binary classifier would not show such a drastic inversion. The magnitude of this skew reflects the model's tendency to make highly confident failure predictions on difficult questions. This calibration property warrants further investigation, particularly before porting selective prediction to deployment settings where confident-failure predictions about specific students could trigger negative downstream consequences.

\FloatBarrier

\subsection{Uncertainty Trajectory Across Question Position}
\label{app:uncertainty_trajectory}
\begin{figure}[ht]
    \begin{minipage}[t]{0.49\linewidth}
        \centering
        \includegraphics[width=\linewidth]{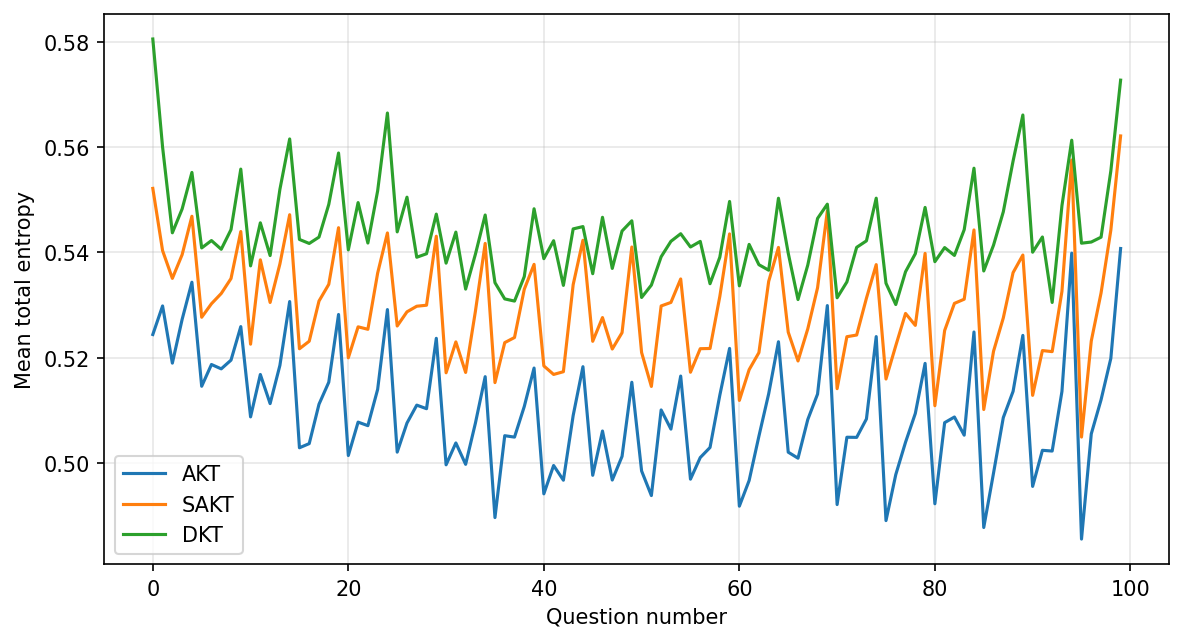}
        \caption{Each model's predictive total entropy against the question number currently being predicted.}
        \label{fig:entropy_vs_qn}
    \end{minipage}
    \hfill
    \begin{minipage}[t]{0.49\linewidth}
        \centering
        \includegraphics[width=\linewidth]{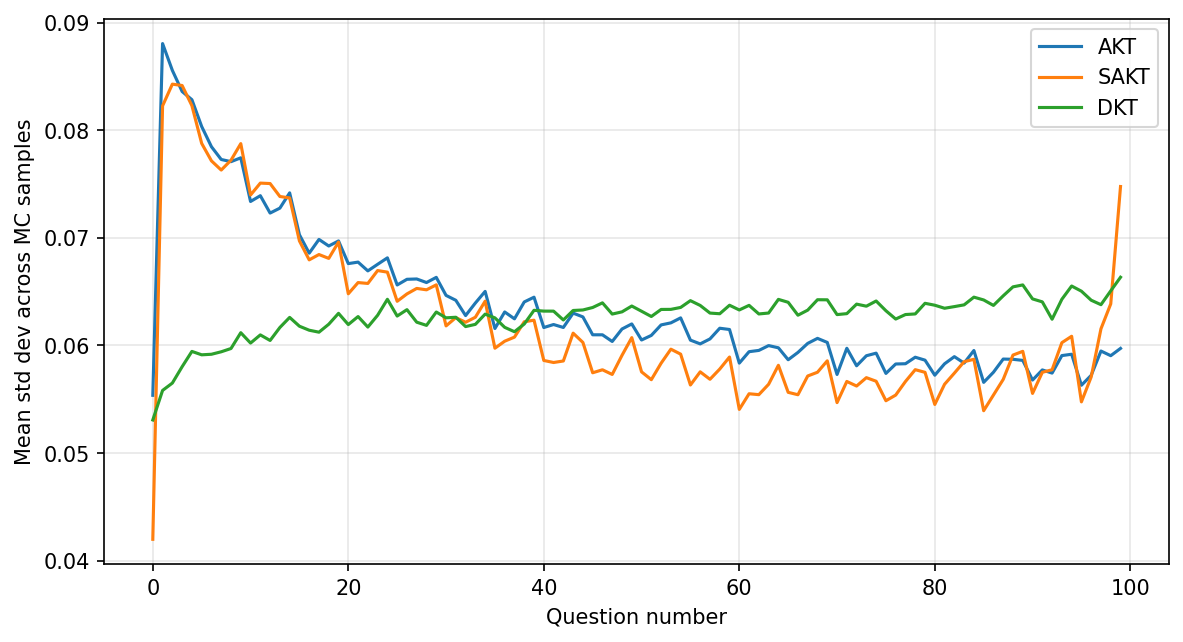}
        \caption{Each model's prediction standard deviation against the question number currently being predicted.}
        \label{fig:std_vs_qn}
    \end{minipage}
\end{figure}

While the mean total entropy remains similar across the different architectures (\Cref{fig:entropy_box_plot}), the mean prediction standard deviation displays visible differences (\Cref{fig:std_box_plot}). DKT exhibits the largest median spread, while SAKT and AKT show similar but smaller spreads. This cross-architecture spread comparison should be interpreted with the caveat that dropout rates differ across models (0.5 for SAKT, 0.2 for AKT and DKT; \Cref{tab:model_hparams}); the observed spread ranking therefore partly reflects this configuration choice rather than purely architectural differences. The broader claim that MC-Dropout exposes architecture-specific epistemic content is robust to dropout-rate calibration, but the per-model spread ordering is not. \Cref{fig:entropy_vs_qn} plots entropy against question position. DKT registers a higher entropy on the very first question than the attention-based models, suggesting that attention-based architectures more easily learn an initial question bias from the training data. A repeating oscillation emerges in the trajectory, periodic with the five-question quiz structure (\Cref{fig:qd_vs_qn} shows the matching question-difficulty oscillation); we examine the entropy--difficulty relationship directly in \Cref{app:uncertainty_difficulty}. \Cref{fig:std_vs_qn} presents the same trajectory using prediction standard deviation. The spread for DKT increases over the first 20 questions before stabilising, whereas SAKT and AKT decrease as more history accumulates. The attention-based models grow more confident over time, while the LSTM-based model does not.

\FloatBarrier

\subsection{Correlation with Question Difficulty}
\label{app:uncertainty_difficulty}
\begin{figure}[ht]
    \begin{minipage}[t]{0.49\linewidth}
        \centering
        \includegraphics[width=\linewidth]{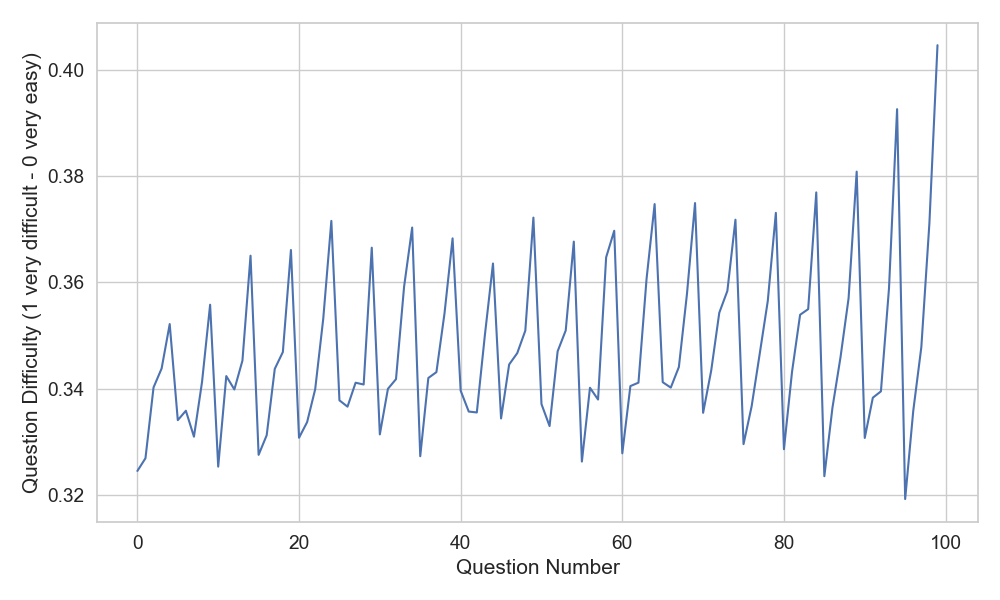}
        \caption{Comparison of question difficulty against the question number that would be predicted by a model.}
        \label{fig:qd_vs_qn}
    \end{minipage}
    \hfill
    \begin{minipage}[t]{0.49\linewidth}
        \centering
        \includegraphics[width=\linewidth]{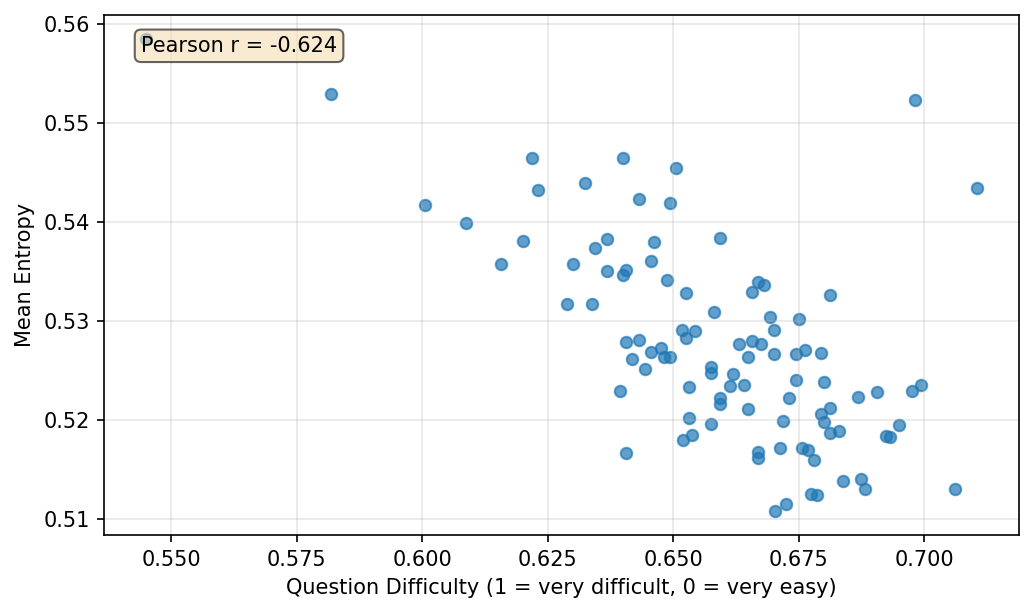}
        \caption{Correlation between the model predictive entropy and the question difficulty.}
        \label{fig:entropy_correlation_qd}
    \end{minipage}
\end{figure}

\Cref{fig:qd_vs_qn} illustrates the mean question difficulty against question position, exhibiting a sawtooth pattern that reflects the within-quiz progression. \Cref{fig:entropy_correlation_qd} reports a moderately strong negative correlation between mean predictive entropy and question difficulty (Pearson $r = -0.62$, $p < 10^{-11}$). The negative sign is a feature of binary correctness prediction under class imbalance: on harder questions the model can confidently predict failure ($p_\text{correct} \to 0$, low entropy), while on easier questions it commits to a moderately confident correct prediction with entropy nearer the binary maximum. The same mechanism produces the slightly higher median entropy on student-correct cases visible in \Cref{fig:entropy_box_plot}; both reflect the same underlying calibration of the binary classifier rather than a failure of the uncertainty signal itself, which remains useful for flagging \emph{model} errors (\Cref{fig:total_entropy}).

\FloatBarrier

\section{Subgroup Fairness Beyond Student Ability}
\label{sec:subgroup_fairness}
\begin{figure}[ht]
  \centering
  \includegraphics[width=\linewidth]{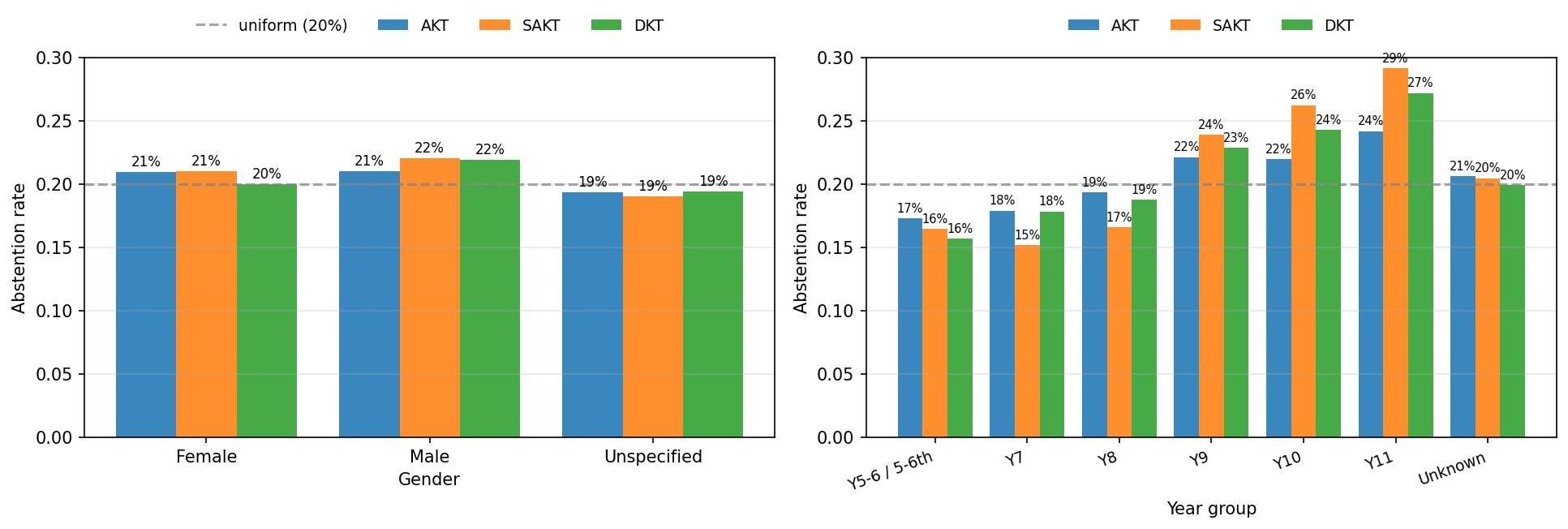}
  \caption{Abstention rate at $c=0.80$ broken down by self-reported gender (left) and country-year group (right). Gender-conditioned rates fall within 19--22\% across all three architectures, very close to the 20\% uniform line. Year-group rates trend upward with student age (15--18\% for Year 5--8 vs.\ 22--29\% for Year 9--11), showing that the abstention is not evenly distributed across cohort age but does not single out any specific group for over-deferral.}
  \label{fig:subgroup_fairness}
\end{figure}

The paper's fairness analysis (\Cref{fig:targeting_fairness}a) shows that abstention rate is close to the 20\% uniform baseline across student-ability quartiles. Eedi's user metadata also records gender and year-group; we extend the fairness check to these subgroups in \Cref{fig:subgroup_fairness}.
\\ \textbf{Gender}: female, male, and unspecified-gender students fall within 19--22\% abstention across all three architectures. The selective layer treats the three groups equivalently within sampling noise.

\textbf{Year group}: abstention rises monotonically with student age, from 15--18\% on Year 5--8 cohorts to 22--29\% on Year 9--11. This trend is consistent across architectures and most likely reflects question difficulty: older students are exposed to objectively harder mathematics content, on which the model has more uncertain predictions. We do not interpret this as a fairness violation, since no specific group is being over-deferred relative to its underlying difficulty profile. However, a deployment system that wished to enforce a strict per-cohort abstention budget would need to apply per-group thresholds rather than a single global $c$. Per-group calibration of the operating point is left to future work.
\FloatBarrier

\section{Variance Decomposition of MC-Dropout Epistemic Uncertainty}
\label{sec:variance_decomp_app}

\paragraph{Question being asked.}
The variance decomposition is a falsification test for the redundancy hypothesis: is MC-Dropout's signal recoverable from publicly observable, model-free features of the (student, question) pair? \\ If yes, MC-Dropout is a complicated re-implementation of cheaper proxies and a deployment system could substitute one for the other; if no, MC-Dropout surfaces signal those proxies cannot reconstruct.

\paragraph{Target variable.}
For each of the $\sim$110{,}000 last-70 prediction targets per architecture, we compute the model's \textit{epistemic} uncertainty using the BALD decomposition~\citep{houlsby2011bayesian}:
\[
\underbrace{H\big[\,\mathbb{E}_w\,p(y\mid x,w)\,\big]}_{\text{total}}
\;-\;
\underbrace{\mathbb{E}_w\,H\big[\,p(y\mid x,w)\,\big]}_{\text{aleatoric}}
\;=\;
\underbrace{I(y;w\mid x)}_{\text{epistemic}},
\]
where the expectation is approximated over $M=100$ stochastic dropout-active forward passes. The aleatoric component is subtracted out so the regression target is purely the model's \textit{reducible} uncertainty over its own learned weights.

\paragraph{The decomposition uses BALD epistemic but the rest of the paper uses total entropy.}
Throughout the paper we use total predictive entropy and prediction variance as the primary uncertainty signals, since these are the quantities the selective-prediction layer actually sorts on and the descriptive figures most directly visualise. The information-theoretic decomposition of total predictive entropy into aleatoric and epistemic components is known to have identifiability and additivity caveats: the additive split does not always correspond to a clean separation of irreducible-versus-reducible uncertainty, and the empirical estimates inherit instabilities from the dropout posterior approximation \citep{wimmer2023quantifying,valdenegro2022deeper}. We therefore avoid leaning on the decomposition where it is not strictly necessary, and use total entropy / variance as the operational signals.

For the variance decomposition specifically, however, the question being asked requires isolating the part of the signal that classical psychometric proxies cannot recover. Proxies like IRT outcome ambiguity ($H(p_{\mathrm{IRT}})$) are aleatoric-flavoured by construction: they only see student ability and question difficulty, never the model's representational state. Regressing total entropy on these proxies would therefore partly be regressing the aleatoric component on its own surrogate, inflating the explained $R^2$ without telling us whether the architecture's epistemic content itself is recoverable. Subtracting the aleatoric estimate via BALD gives a more honest test, even if the subtraction is imperfect. As a robustness check, we re-ran the linear decomposition with total predictive entropy as the regression target and found the qualitative conclusion intact: classical psychometric proxies still leave 60\%--64\% of the signal unexplained across all three architectures. After the aleatoric component is subtracted out, the same linear regression on BALD epistemic uncertainty leaves 96\%--99\% unexplained, falling to 77\%--90\% only when we permit a non-linear regressor (see below). The architecture-specific dominance therefore does not depend on either the BALD subtraction or the linear functional form. It is amplified by both, but already present without them.

\paragraph{Predictors.}
Each prediction is annotated with classical psychometric features that a deployment system could compute without running the model:
\begin{itemize}\setlength\itemsep{0.2em}
  \item \textbf{Question difficulty}: empirical correctness rate of question $q$ across the validation cohort.
  \item \textbf{Student ability}: per-user accuracy across that user's full 100-question sequence.
  \item \textbf{IRT outcome ambiguity}: $H(p_{\mathrm{IRT}})$, the Shannon entropy of the 2PL prediction $p_{\mathrm{IRT}}(u, q) = \sigma(a_q(\theta_u - b_q))$, with $\theta_u$ fit on the first 30 questions.
  \item \textbf{Curriculum coverage}: four binary indicators (one per curriculum level) for whether the target question's subject / topic / subtopic / construct appears in the same student's first-30 context.
\end{itemize}

\paragraph{Procedure.}
We fit a sequence of ordinary least-squares linear regressions of epistemic uncertainty on \emph{nested} predictor sets, computing the coefficient of determination $R^2$ at each step. The marginal $R^2$ of the $k$-th predictor is the additional fraction of variance in epistemic uncertainty linearly explained by adding that predictor on top of those already in the model; the cumulative $R^2$ at step $k$ is the total fraction jointly explained by the first $k$ predictors. The unexplained residual $1 - R^2_{\text{full}}$ is therefore signal that no linear combination of these proxies captures.

\begin{figure}[ht]
  \centering
  \includegraphics[width=0.95\linewidth]{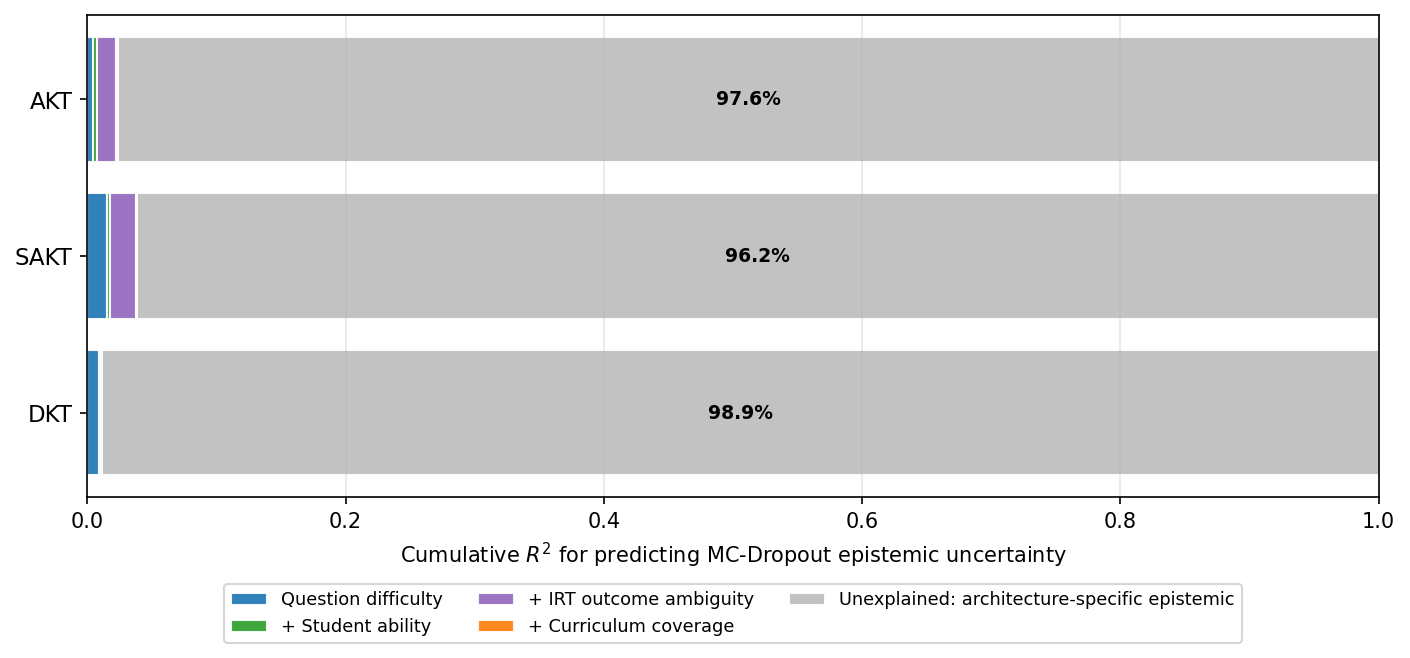}
  \caption{Variance decomposition of MC-Dropout epistemic uncertainty (BALD), under linear modeling. Cumulative $R^2$ from question difficulty, student ability, IRT outcome ambiguity, and curriculum coverage at all four granularities tops out at 3.8\% across all three architectures. A non-linear robustness check (5-fold cross-validation random forest, see text) raises explained $R^2$ to 9.8--23.2\%, leaving 76.8--90.2\% as architecture-specific content that classical psychometric proxies cannot recover even at a much richer model class.}
  \label{fig:what_drives}
\end{figure}

\paragraph{Interpretation of the result.}
\Cref{fig:what_drives} reports cumulative $R^2$ at each step, for each architecture. The full classical psychometric stack jointly explains 2.4\% (AKT), 3.8\% (SAKT), and 1.1\% (DKT) of epistemic uncertainty under linear modeling. Because each marginal contribution is small, no single predictor is doing the bulk of the work either: this is not the case where one proxy already captures everything and the others add nothing redundant.

\paragraph{Non-linear robustness check.}
A reviewer might object that linear regression is too conservative: perhaps the same predictors, combined non-linearly, capture much more. To bound this we re-fit the predictor stack as input to a 5-fold cross-validated random forest (\texttt{n\_estimators=100}, \texttt{min\_samples\_leaf=20}); this is a much more flexible model class but uses identical features and the cross-validation prevents overfitting from inflating $R^2$. Cross-validated $R^2$ values are 10.5\% (AKT), 9.8\% (SAKT), and 23.2\% (DKT), leaving 89.5\%, 90.2\%, and 76.8\% unexplained respectively. The non-linear gap-closing is largest for DKT (the LSTM-based architecture), suggesting its epistemic uncertainty has more recoverable interactional structure than the attention-based architectures, but a substantial majority of the signal remains architecture-specific in every case. The model's epistemic uncertainty therefore varies on dimensions that are not just linearly orthogonal to the classical psychometric features, but largely orthogonal at any reasonable model class: that variation is what MC-Dropout's stochastic forward passes recover.

\FloatBarrier

\end{document}